# Rapid head-pose detection for automated slice prescription of fetal-brain MRI

Malte Hoffmann[1,2] | Esra Abaci Turk[3,4] | Borjan Gagoski[2,3] |
Leah Morgan[1] | Paul Wighton[1] | Matthew Dylan Tisdall[5] |
Martin Reuter[1,2,6] | Elfar Adalsteinsson[4,7] | Patricia Ellen Grant[2,3] |
Lawrence L. Wald[1,2] | André J. W. van der Kouwe[1,2]

[1]Department of Radiology, Massachusetts General Hospital, Boston, Massachusetts

[2]Department of Radiology, Harvard Medical School, Boston, Massachusetts

[3]Fetal-Neonatal Neuroimaging and Developmental Science Center, Boston Children's Hospital, Boston, Massachusetts

[4]Electrical Engineering and Computer Science, Massachusetts Institute of Technology, Cambridge, Massachusetts

[5]Radiology, Perelman School of Medicine, University of Pennsylvania, Philadelphia, Pennsylvania

[6]German Center for Neurodegenerative Diseases, Bonn, Germany

[7]Institute for Medical Engineering and Science, Massachusetts Institute of Technology, Cambridge, Massachusetts

**Correspondence**
Malte Hoffmann, Department of Radiology, Massachusetts General Hospital, Boston, MA, USA.
Email: mhoffmann@mgh.harvard.edu

**Funding information**
National Institute of Biomedical Imaging and Bioengineering, Grant/Award Number: R01-EB008547; National Institute of Child Health and Human Development, Grant/Award Numbers: K99-HD101553, R00-HD074649, R01-HD085813; National Institute of Neurological Disorders and Stroke, Grant/Award Number: U01-AG052564

**Abstract**

In fetal-brain MRI, head-pose changes between prescription and acquisition present a challenge to obtaining the standard sagittal, coronal and axial views essential to clinical assessment. As motion limits acquisitions to thick slices that preclude retrospective resampling, technologists repeat ~55-second stack-of-slices scans (HASTE) with incrementally reoriented field of view numerous times, deducing the head pose from previous stacks. To address this inefficient workflow, we propose a robust head-pose detection algorithm using full-uterus scout scans (EPI) which take ~5 seconds to acquire. Our ~2-second procedure automatically locates the fetal brain and eyes, which we derive from maximally stable extremal regions (MSERs). The success rate of the method exceeds 94% in the third trimester, outperforming a trained technologist by up to 20%. The pipeline may be used to automatically orient the anatomical sequence, removing the need to estimate the head pose from 2D views and reducing delays during which motion can occur.

## 1 | INTRODUCTION

Fetal-brain magnetic resonance imaging (MRI) has become an invaluable tool for understanding early brain development and can remove diagnostic doubt that may remain after routine ultrasound exams.[1,2] Unfortunately, fetal and maternal motion limit fetal MRI to rapid two-dimensional (2D) sequences such as half-Fourier single-shot turbo spin-echo (HASTE).[3,4] Although HASTE decreases the impact of motion during the sub-second acquisition of an individual slice, multiple slices are needed, and acquiring consecutive orthogonal scans is challenging as motion occurs between slices and between scans.[5]

### 1.1 | Chasing the fetus

In particular, head-pose changes between prescription and acquisition frequently produce images with incorrect anatomical orientation as shown in Figure 1A. However, non-oblique views of the standard sagittal, coronal or axial planes are essential for clinical assessment.[6,7]

Incorrect anatomical orientation does not pose a problem to MRI applications that acquire isotropic 3D data, since image volumes can be resampled after the fact, provided motion does not occur during the 3D acquisition. In fetal 2D MRI, this is impractical because the slice thickness typically exceeds the ~1-mm in-plane voxel dimension by at least threefold, and data across slices may be inconsistent due to motion and spin-history effects,[8] as can be seen in Figure 1B.

As a result, fetal MRI sessions are dominated by the inefficient practice of "chasing the fetus." This typical workflow consists of repeating the ~55-s stack-of-slices acquisition until motion-free volumes are obtained in all three planes (see Figure 1C).

### 1.2 | Background

In conventional MR applications, approaches to automating the scan prescription are successfully used to remove user-dependent variability. Automated slice positioning in adult-brain MRI[9,10] and single-voxel placement for MR spectroscopy[11-13] remove the need for manual positioning of the field of view (FOV) using a quick pre-scan. This "scout" acquisition is run before the actual MR sequence and registered[14,15] to an atlas image to estimate the orientation of the brain relative to the scanner coordinate system. Unfortunately, unsupervised registration is challenging in fetal MRI due to signal from amniotic fluid and maternal tissue surrounding the brain,[16] despite the availability of atlases that capture the development of the brain at different stages of gestation.[17]

Recent developments focus on fetal-brain extraction and segmentation for post-processing applications, in full ~55-s structural MRI data with high in-plane resolution. An overview is provided in the following sections.

#### 1.2.1 | Image-processing techniques

Template matching is a processing technique for finding areas in a target image that are similar to a template. Typically, a sliding window is moved over the target and a similarity metric is evaluated to assess how well each window position matches the template.[18,19] Anquez et al

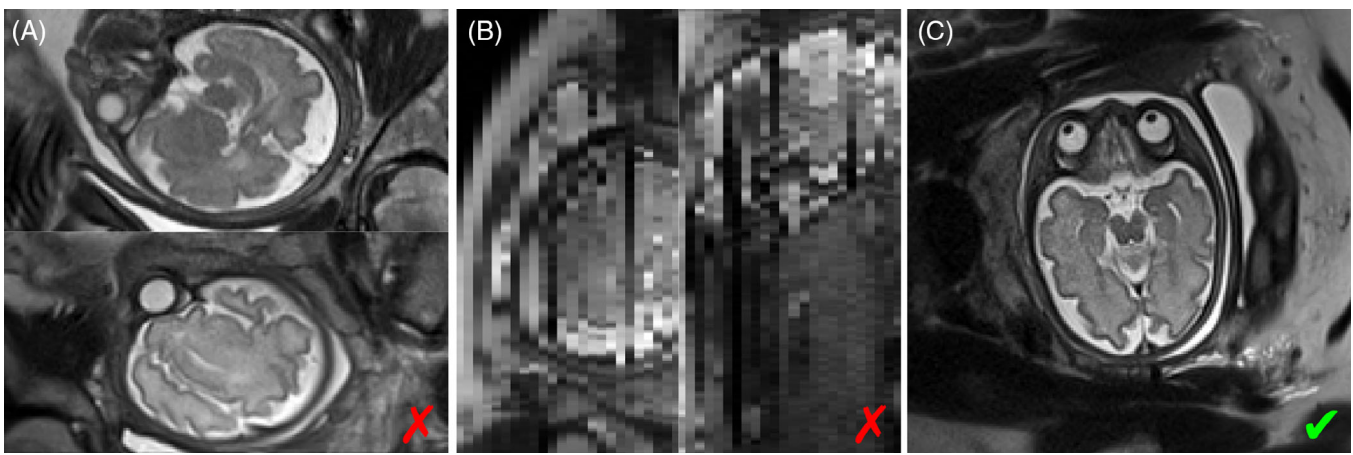

**FIGURE 1** (A) Half-Fourier single-shot turbo spin-echo (HASTE) slices with incorrect anatomical orientation due to motion between prescription and acquisition. (B) Resampling across slices, as shown here for the bottom image in (A), is not viable due to inconsistent data across slices and slice thicknesses that typically exceed the in-plane voxel dimension by threefold. (C) Near-perfect axial slice. All images were acquired from the same fetus

demonstrate automated brain extraction from anatomical fetal MRI in ~5 min using local graph cuts after template-matching based localization of the fetal eyes.[18] Another method uses morphological operations and connected component analysis to isolate the brain from single HASTE slices known to include the fetal brain with higher intensity than the surrounding tissues.[20]

### 1.2.2 | Feature-based classification

A class of machine-learning approaches trains classification algorithms to distinguish between sets of relevant features extracted from the image.[21-23] One method transforms the fetal-MRI scans into superpixel graphs to train a random-forest classifier on scale-invariant feature transform (SIFT)[24] features.[21] Keraudren et al filter maximally stable extremal regions (MSERs)[25] by shape and size before random-forest classification using bundled SIFT features.[23] These approaches achieve high brain-localization and segmentation accuracies in sub-millimeter fetal MRI, typically requiring 3–5 min.

### 1.2.3 | Deep-learning approaches

Sub-second runtimes are achieved by deep convolutional neural networks (CNNs), which remove the need for separate feature extraction.[26-29] Segmentation CNNs build on a U-Net architecture[30] and train on relatively large datasets, although augmentation strategies have been used to expose models to more variability than the training data encompasses.[31,32] For many fetal applications, training requires manual segmentations[27] or other ground-truth information[33] to be available.

Recent advances employ convolutional encoders to estimate the fetal-head pose from HASTE-slice stacks and orient individual image slices within a template volume.[33,34] For accurate prediction, these networks require input data with pre-segmented fetal brains and N4 bias-field correction at test time, which can take ~10 min per image volume.[35]

### 1.3 | Motivation

A common theme of recent work is the focus on retrospective analysis and processing of high-resolution data. A promising application of such methods is super-resolution reconstruction (SVR), which produces image volumes with isotropic resolution from a minimum of nine HASTE stacks acquired with different orientations relative to the fetal brain.[36-38] This can lead to excellent image quality if multiple input stacks with good anatomical coverage and high in-plane quality are available,[36,37] but the end result cannot be assessed before acquiring several slice stacks and completing the ~10-min reconstruction.[37]

In contrast to SVR, our ultimate goal is to avoid redundant sampling and automate the acquisition pipeline to acquire only the data needed by radiologists, available for immediate interpretation. Immediate availability is crucial in the clinic as pregnant women and their fetuses are a particularly vulnerable population: at our hospital and many others, a radiologist is present during fetal-MRI sessions to assess the diagnostic quality of acquisitions and guide the technologist in changing the protocol as needed.[39-41] This deviates from the conventional radiological practice of having the technologist steer the session and interpreting images at a later time, where follow-up exams are deemed acceptable in the rare event that more anatomical detail or better image quality are needed.

We focus on fetal-head pose detection in low-resolution scout images (echo planar imaging, EPI), which can be acquired in seconds and used to automatically prescribe HASTE slices with the desired orientation. However, such a general-purpose approach will not be limited to HASTE. It could be used to prescribe *any* pulse sequence when prepended with the scout scan, and fast automated prescription will reduce the vulnerability of fetal-brain MRI to pose changes, which will hopefully improve patient comfort by reducing the duration of fetal-MRI sessions. In addition, the approach may also benefit SVR-specific data acquisitions, for example, by automatically ensuring sufficient angular and anatomical coverage or shortening the acquisition time of individual stacks by tailoring the number of slices to the spatial extent of the brain. The remaining two axes of the FOV have less potential for optimization: while oversampling in readout direction incurs no time penalty, the FOV in phase-encode direction has to fully encompass the maternal anatomy to avoid wrap-around artifacts.

Crucially, the head-pose detection needs to be efficient and cannot rely on extensive pre-processing if it is to be integrated into the acquisition pipeline. Our motivation for using traditional image processing is full control over each stage of the algorithm, such that individual components can be optimized for real-time use on the scanner. Current clinical systems have no GPUs or, as our 3-T research system (MAGNETOM Prisma, Siemens Healthineers, Erlangen, Germany), older-generation GPUs such as a Tesla K10 with 1.8 GB of memory (Nvidia, Santa Clara, CA), forbidding real-time use of U-Nets if the approach is to be run within the reconstruction pipeline.

## 1.4 | Contribution

In this work, we propose a robust fetal-head pose detection pipeline for full-uterus low-resolution EPI scans with limited contrast. We leverage an efficient MSER detector[42] to achieve runtimes of ~2 s. Similar to Anquez et al, we identify the fetal eyes as landmarks but we begin with a global search of the brain to improve robustness to limited contrast.[18] To increase efficiency, we avoid template matching. We build on the idea of filtering MSERs using prior knowledge of the size and shape of the brain as a function of age,[23] and extend it to the eyes to achieve robustness across a range of fetal development. However, we do not need to rely on classification algorithms.

Our contribution includes the following points. We present automatic fetal brain and eye detection tailored for fast full-uterus low-resolution EPI stacks, that can be acquired in seconds as a scout scan. The algorithm includes a rapid segmentation step that creates binary 3D masks for the brain and eyes by combining and regularizing MSERs across image slices. We propose a fetal head-pose detection in 3D space based on these landmarks. We estimate the transformation from image to anatomical space, which could be used to automatically prescribe a subsequent scan in standard anatomical planes. We quantify the prevalence of fetal head-pose changes in the clinic by determining the success rate of a trained technologist prescribing HASTE slices on the scanner. We evaluate the performance of our method in EPI across a range of contrasts, fetal age, abnormalities and motion levels. We measure the accuracy of the brain location and orientation as compared to a trained labeler.

## 2 | METHOD

The problem addressed in this work is incorrect anatomical orientation of structural fetal-brain HASTE. While we focus on processing EPI scout acquisitions to enable automatic prescription of a subsequent scan, we first survey clinical HASTE stacks to quantify the prevalence of the problem in the clinic.

Data specific to this work are acquired using an IRB-approved protocol, and written informed consent is obtained from each participant before the experiment. Separate IRB approval is obtained for the retrospective analysis of data from our clinical archive.

## 2.1 | Problem quantification: clinical prescription success rate

We rate randomly selected clinical HASTE stacks from our picture archiving and communication system (PACS) to quantify the extent to which incorrect anatomical orientation corrupts fetal MRI, that is, the prescription success rate of a trained technologist. In contrast to motion during the scan, our focus is on prescription errors and pose-changes between the prescription and the acquisition. For example, if the first 30% of slices are acquired with correct anatomical orientation but all later slices are corrupted by subject motion, we will regard the prescription as successful whereas a radiologist might consider the stack as undiagnostic.

All images were acquired at Boston Children's Hospital between July 2013 and March 2018. We focus on neurologically indicated exams and disregard full-uterus "overview" stacks obtained for planning. We classify stacks as "success" or "failure" based on their orientation with the anatomical axes, permitting in-plane rotations: scans with axial and coronal slices are deemed successful if the midline of the brain remains stationary while scrolling through slices, whereas scans with sagittal slices are considered successful if both eyes appear in the same in-plane location when scrolling through slices. Scans are solely evaluated based on in-plane views.

If in doubt, no label is assigned, and we rate complete sessions until more than 100 HASTE stacks are labeled. Overall, we rate 106 stacks acquired in 13 sessions from 10 mothers (two have more than one session). A second blinded rater repeats the procedure to estimate inter-rater reliability.

We also quantify the typical duration and delay between fetal-brain acquisitions by parsing the DICOM headers of the same data. This analysis considers only immediately successive HASTE stacks.

## 2.2 | Proposed method overview

We propose to derive the current left–right, posterior–anterior and inferior–superior axes of the fetal brain from a single EPI volume, which will enable automatic prescription of a subsequent scan (e.g., HASTE) with its FOV aligned with these axes.

Our algorithm finds the location of the fetal brain and eyes by detecting MSERs on each slice of the image and combining them in 3D space. These landmarks, in combination with the geometry of a 3D brain mask created during the process, provide sufficient information to fully determine the orientation of the brain. Figure 2 gives a high-level overview of our method.

As MSERs can be detected in tissues other than the brain or eyes, we leverage filtering procedures and prior knowledge of anatomical size and shape to reject irrelevant regions. Gestational age (GA) is an *input* to our algorithm that accommodates detecting appropriately sized

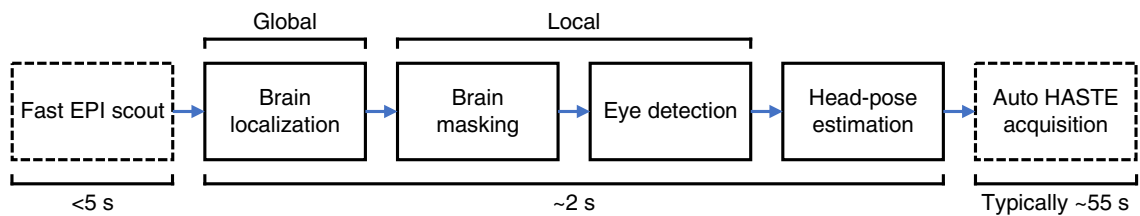

**FIGURE 2** Fetal-brain localization and head-pose detection. After the brain is localized in a quick scout scan (global), subsequent steps are performed in a subset of the image centered on the brain (local). Fast head-pose detection will enable automated slice prescription with half-Fourier single-shot turbo spin-echo (HASTE) and is to be embedded on-scanner as illustrated by the boxes drawn with dashed lines. However, this general-purpose approach could be used to prescribe *any* pulse sequence

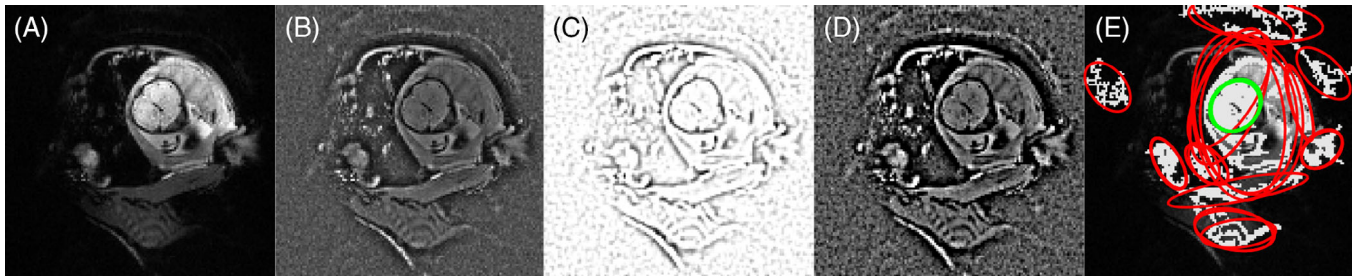

**FIGURE 3** Real-time image preparation and feature detection. (A) Echo planar imaging (EPI) slice of the pregnant abdomen showing the fetal brain. (B) To improve the detection of maximally stable extremal regions (MSERs), image intensities are flattened. (C) We enhance edges in the image by multiplying voxel intensities by weights derived using a difference-of-Gaussian filter (image shows final weights). (D) Final image used for MSER detection. (E) MSERs (white) likely to contain the brain (green) are delineated from non-brain MSERs (red) by fitting ellipses and filtering these based on prior anatomical knowledge

MSERs for robust orientation estimation across a range of GA. This is detailed in the following sections.

### 2.2.1 | Real-time image preparation

To facilitate MSER detection, image intensities are flattened by dividing each slice by a blurred version of itself, using a Gaussian kernel with isotropic full width at half maximum (FWHM) $f_{NU} = 20$ mm. Dark edges are enhanced by weighting voxels based on a difference-of-Gaussian (DOG) filter[43] with isotropic FWHM $f_1 = 10$ mm and $f_2 = 5$ mm as follows: voxel weightings are computed by applying the DOG filter to a copy of the image, setting negative voxels to zero, rescaling into the interval [0, 1] and inverting intensities. Figure 3A–D illustrates these steps for a slice of a typical EPI volume.

### 2.2.2 | Slice-wise MSER detection

MSERs are connected components of the level sets of an image that are stable in size across a range of intensities. They can be detected as follows. All pixels are removed from the image and sorted by intensity. We step through intensity levels and re-insert the pixels of each level. As pixels are added, clusters appear in the image and begin to grow. Tracking the growth of a cluster across intensity levels, an MSER corresponds to its pixels at a local minimum of the growth curve. MSERs are well suited for our purpose since cerebrospinal fluid, brain and eyes in T2-weighted fetal MRI are brighter than the surrounding tissues.[23] We implement an efficient algorithm for MSER detection[42] and extract these regions from each slice of the image.

## 2.3 | Brain localization

### 2.3.1 | Fetal age-specific MSER filtering

We fit an ellipse to the binary mask of each MSER to measure its basic geometric properties.[23] This enables us to reject a large proportion of non-brain MSERs based on prior knowledge of the occipitofrontal diameter (OFD) and biparietal diameter (BPD) at corresponding GA (data available for 14–40 weeks).[44] Specifically, we restrict the major and minor ellipse axes $a$ and $b$, ellipse area $A$, aspect ratio $r$ and the Dice overlap coefficient $D$ between

**TABLE 1** Restrictions on accepted MSERs

| Geometric property | (A) Brain localization | | (B) Brain masking | | (C) Eye detection | |
| --- | --- | --- | --- | --- | --- | --- |
| | Min | Max | Min | Max | Min | Max |
| $A\left(\pi \cdot \frac{a_{ref}}{2} \cdot \frac{b_{ref}}{2}\right)^{-1}$ | 0.2 | 1.1 | 0.05 | 1.1 | 0.5 | 1.2 |
| $r\left(\frac{a_{ref}}{b_{ref}}\right)^{-1}$ | – | 1.5 | – | 1.5 | – | 1.5 |
| $a\, a_{ref}^{-1}$ | – | 1.1 | – | 1.2 | – | 1.3 |
| $D$ | 0.7 | – | 0.7 | – | 0.9 | – |
| $d_{brain}\left(\frac{OFD}{2}\right)^{-1}$ | – | – | – | 1.1 | 0.6 | 1.3 |
| $C$ | – | – | – | – | 2 | – |

*Note:* The filtered properties include fitted ellipse area $A$, aspect ratio $r$, major-axis length $a$, ellipse-MSER Dice overlap $D$ and distance $d_{brain}$ from the brain center. For the brain, occipitofrontal ($a_{ref}$ = OFD) and biparietal diameter ($b_{ref}$ = BPD) are used as reference values. For the eyes, ocular diameter and local contrast $C$ are considered instead ($a_{ref} = b_{ref}$ = OD). OFD, BPD and OD are available and used for filtering as a function of fetal age (14–40 weeks) for robustness across different stages of gestation.

Abbreviations: BPD, biparietal diameter; MSERs, maximally stable extremal regions; OD, ocular diameter; OFD, occipitofrontal diameter.

the MSER mask and fitted ellipse. Only MSERs matching the constraints in Table 1A are retained. Examples of typical brain and non-brain MSERs are shown in Figure 3E.

### 2.3.2 | Clustering across slices

We use nonparametric density gradient estimation[45] to cluster MSERs across slices. Briefly, a regular grid of starting points $P$ is laid over the FOV in 3D. At each iteration, each point is moved to a new position

$$P'(P) = \frac{\sum_{i=1}^{n} K(\|x_i - P\|) x_i}{\sum_{i=1}^{n} K(\|x_i - P\|)}$$

where $x_i$ ($i \in \{1, 2, ..., n\}$) is the barycenter of MSER $i$ and $K$ is a Gaussian kernel of FWHM $f_{clust} = 0.5 \times$ OFD. After the set of points converges to the $m$ centroids $C_j$ ($j \in \{1, 2, ..., m\}$) of clusters with the highest spatial density of MSERs, all MSERs $i$ within the expected spatial scale of the fetal brain are considered part of cluster $j$, that is, if they satisfy the condition

$$\|x_i - C_j\| < 0.5 \times \text{OFD}$$

We identify the brain by selecting the cluster with the largest binary 3D mask $M_j$, encouraging near-ellipsoidal shape, where $M_j$ is obtained by aggregating the voxels of all MSERs associated with the cluster. This is done by ranking clusters according to a simple fill factor

$$F_j = |M_j \cap E_j| - |E_j \setminus M_j|$$

where $E_j$ designates the set of voxels encompassed by an ellipsoid fitted to the 3D mask $M_j$ and the number of voxels is denoted by $|\cdot|$. There is no restriction on how close to spherical the ellipsoid can be. The brain center $B$ is estimated from the mask $M$ of the cluster with the highest fill factor as follows.

### 2.3.3 | Fetal-brain center

We add voxels to the mask $M$ to account for MSERs that were rejected or not detected on intermediate image slices. First, we morphologically fill any holes within each slice of the mask $M$. Second, empty slices are filled in by adding voxels that are part of $M$ on both of the closest non-empty slices on each side. Third, we remove protrusions on each slice by taking away any voxel not included on the closest slice on at least one side. Finally, the brain center $B$ is taken to be the barycenter of $M$.

## 2.4 | Brain masking

In the brain localization procedure, the size of MSERs is restricted to improve the computational efficiency and robustness of the global search. Once the location of the brain is known, we perform a more exhaustive local search, with the goal of constructing a binary mask that encompasses the entire fetal brain.

### 2.4.1 | Local search

We detect MSERs on partial image slices within a box of side length $l = \text{sqrt}(2) \times$ OFD centered on the brain.

Possible non-brain MSERs are rejected as described Section 2.3.1, using the constraints listed in Table 1B, now additionally excluding MSERs based on their distance $d_{brain}^i = \|x_i - B\|$ from the brain. The voxels of all MSERs remaining are combined with the binary mask $M$ from Section 2.3.

In the next section, this mask is regularized to remove marginal non-brain slices, for example, due to MSERs from amniotic fluid detected between the fetal brain and the edge of the amniotic sac. This is achieved by requiring the cross section of the mask to increase while stepping through slices, up to a maximum, and to decrease thereafter.

### 2.4.2 | Spatial regularization

First, we estimate ellipse properties for each slice of the brain mask $M$ and fit a parabola to the major axes of the central 50% of slices, which are usually the largest. Second, we convert the deviation of measured axis lengths from the fit to z-scores and discard any mask slice with z exceeding $z_{poly} = 1.5$ (derived empirically, see below) as illustrated in Figure 4. This is only done if the parabola is concave to prevent rejecting slices in case of a bad fit. Third, gaps resulting from the regularization are closed using 1D morphological hole filling in the slice direction, and protruding voxels are removed (see Section 2.3.3).

Finally, we fit an ellipsoid to the brain mask (in 3D). Its center is taken as an updated estimate of $B$, whereas its major axis $a$ will be useful later for resolving the left–right ambiguity in determining the fetal-brain geometry.

## 2.5 | Eye detection

The search for the fetal eyes can be restricted spatially since the brain location is known. Thus, we only consider partial slices within the box of Section 2.4.1, resampled with halved voxel dimensions to facilitate the detection of smaller structures. MSERs are detected on each slice of the box as detailed above.

### 2.5.1 | Fetal age-specific MSER filtering

As with the brain, we only retain features that are broadly compatible with the anatomical scales expected at the specific age of the fetus[46] to promote robustness across different stages of gestation. Filtering criteria are summarized in Table 1C, including local contrast $C$,

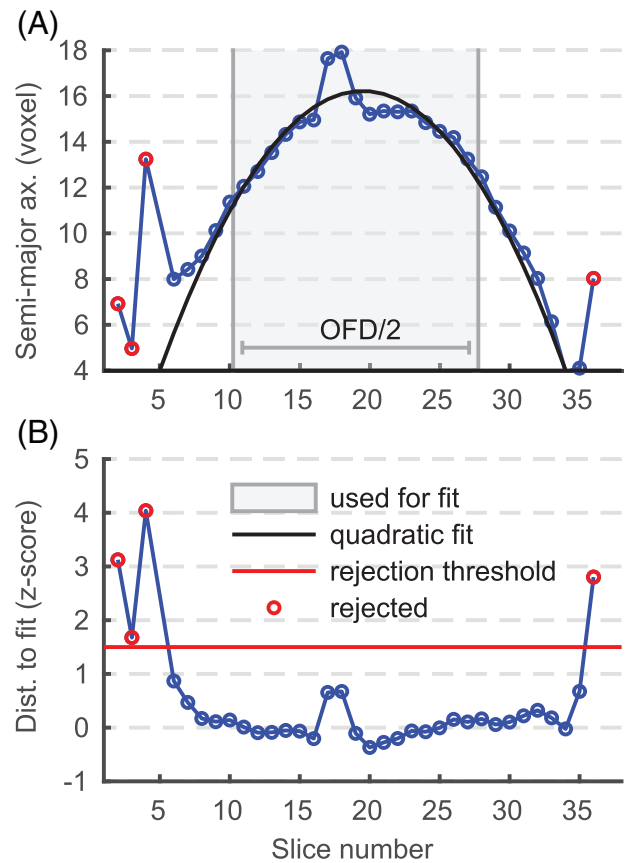

**FIGURE 4** Spatial brain-mask regularization. (A) To remove marginal non-brain slices, a parabola is fitted to the major axes of the central 50% of brain-mask slices, that is, typically the largest slices (shaded area). (B) Normalized deviation of measured axis lengths from the fit. Slices with z-scores exceeding $z_{poly} = 1.5$ (red line, derived empirically) are discarded as indicated by red circles

which measures the extent to which the eyes appear brighter than the surrounding tissues.

For each MSER $i$, local contrast $C_i$ is calculated relative to a ribbon $R_i = \tilde{E}_i \setminus E_i$ of voxels enclosing the ellipse $E_i$ fitted to $i$, where the ellipse $\tilde{E}_i$ is identical to $E_i$ except that its axes are 50% longer. Denoting the median intensity across a set of voxels by $\bar{I}(\cdot)$, we compute local contrast as

$$C_i = \frac{\bar{I}(E_i)}{\bar{I}(R_i)}$$

### 2.5.2 | Clustering across slices

We identify clusters of MSERs across slices based on prior knowledge of the ocular diameter (OD) as a function of fetal age, to accommodate fetal development.[46]

Each MSER, combined with all other MSERs within a radius $\varrho_{\text{clust}} = 0.75 \times \text{OD}$, is considered a potential eye. This approach typically yields $m = 10\text{–}20$ unique clusters, some of which will result from the same structures, and helps identify an optimal anatomical configuration in the presence of inter-slice motion.

To find the two clusters that represent the eyes, we compute the deviation $\epsilon$ from the expected geometry for each pair of clusters $(j, k) \in \{1, 2, ..., m\}^2$ and select the pair with the lowest value:

$$\epsilon(j,k) = \left| \frac{\overline{BM_j} + \overline{BM_k}}{\text{OFD}} - 1 \right| + 2\lambda$$

$$\cdot \frac{|\overline{BM_j} - \overline{BM_k}|}{\overline{BM_j} + \overline{BM_k}} + \left| \frac{\overline{M_j M_k}}{\text{OD}} - 1 \right| + \left| \frac{s_j + s_k}{2 \max_p s_p} - 1 \right|$$

where $B$ is the brain center, $M_j$ is the barycenter of cluster $j$, $\lambda = 4$ is a weight (determined empirically, see below), $s_j$ counts the number of slices cluster $j$ has MSERs on and $\overline{AB}$ denotes the distance between points $A$ and $B$. The first term of $\epsilon$ compares the distance between brain and eyes to prior knowledge, whereas the second term encourages this distance to be similar for both eyes. The third term regularizes the spatial separation between both eyes, and the fourth term promotes clusters with MSERs detected on several image slices.

Once the most likely location of the eyes is determined, we can reconstruct the orientation of the brain in space.

## 2.6 | Fetal-brain orientation

The anatomical axes are fully determined by the geometry of the brain mask $M$ and the barycenters $E_{1,2}$ of the eyes. In particular, any left–right ambiguity can be resolved considering the major-axis $a$ of $M$.

The left–right axis $\text{LR} = E_1 E_2 / \|E_1 E_2\|$ passes through both eyes. We construct the posterior–anterior axis PA from the brain center $B$ through the midpoint between the eyes. The inferior–superior axis IS is obtained by perpendicularity to LR and PA. However, due to motion during the acquisition, LR and PA may not be strictly perpendicular. An orthonormal basis is obtained by replacing LR with the cross product of PA and IS, as is illustrated in Figure 5. Since the plane defined by $\{B, E_1, E_2\}$ is usually tilted from the axial plane by $\theta \approx 30°$ about LR, we replace the anatomical basis {LR, PA, IS} with $R_{\text{LR}}(\theta)$ (LR PA IS), where $R_{\text{LR}}$ is a rotation matrix about LR.

Without further knowledge, it would be impossible to differentiate between the left and right eye. As a result, LR would be flipped half of the time, causing IS to point from the superior to the inferior half of the brain. However, this ambiguity can be fully resolved by considering the major axis $a$ of the ellipsoid previously fitted to the brain mask $M$ as shown in Figure 6: the plane $H$ that contains $a$ and is parallel to LR divides the brain into inferior and superior portions—the inferior part lies in the half space containing the eyes.

## 2.7 | Validation

### 2.7.1 | Data

A total of $n = 41$ full-uterus stacks of T2*-weighted slices from 18 different fetuses in the third trimester (26–37 weeks GA, mean $\pm$ SD 31.0 $\pm$ 2.7) with varying oblique and double-oblique FOV relative to the brain are used for evaluation. The acquisition plane with respect to the mother is arbitrary, since some of the subjects are scanned in the supine and some in the lateral position. Sequence parameters are varied within a range of values to assess robustness across EPI contrasts. We use interleaved single-shot 2D GE-EPI with $(3 \text{ mm})^2$ in-plane resolution, 30–90 3-mm slices (mean 64) and 0–3 mm gap: matrix size $(110\text{–}160)^2$, repetition/acquisition time (TR/TA) 2000–7240 ms, echo time (TE) 28–80 ms, bandwidth 2085–2405 Hz/px, flip angle (FA) 70°–90°, acceleration (GRAPPA) 2, no partial Fourier.

All EPI stacks are acquired on a 3-T Skyra system (Siemens Healthineers, Erlangen, Germany), and written

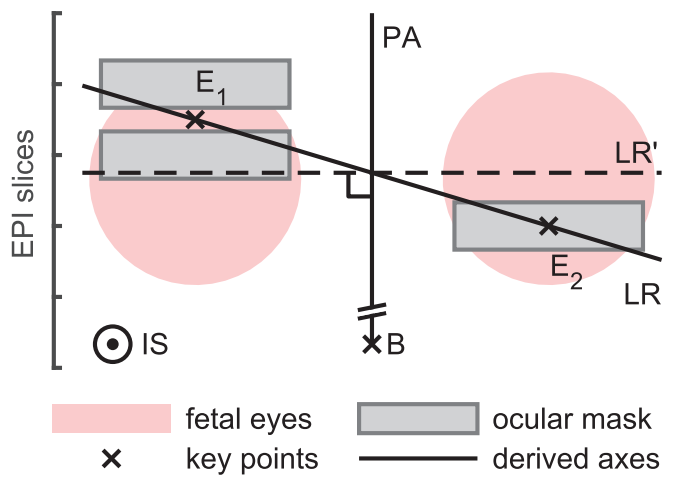

**FIGURE 5** Orthonormal basis construction. The inferior–superior axis IS is constructed by perpendicularity to the left–right (LR) and posterior–anterior (PA) axes. Due to imperfect masking and motion, LR and PA may not be strictly perpendicular. This can be resolved by replacing LR with the cross product $\text{LR}' = \text{PA} \times \text{LA}$, although LR' may not pass through the eyes as accurately as LR

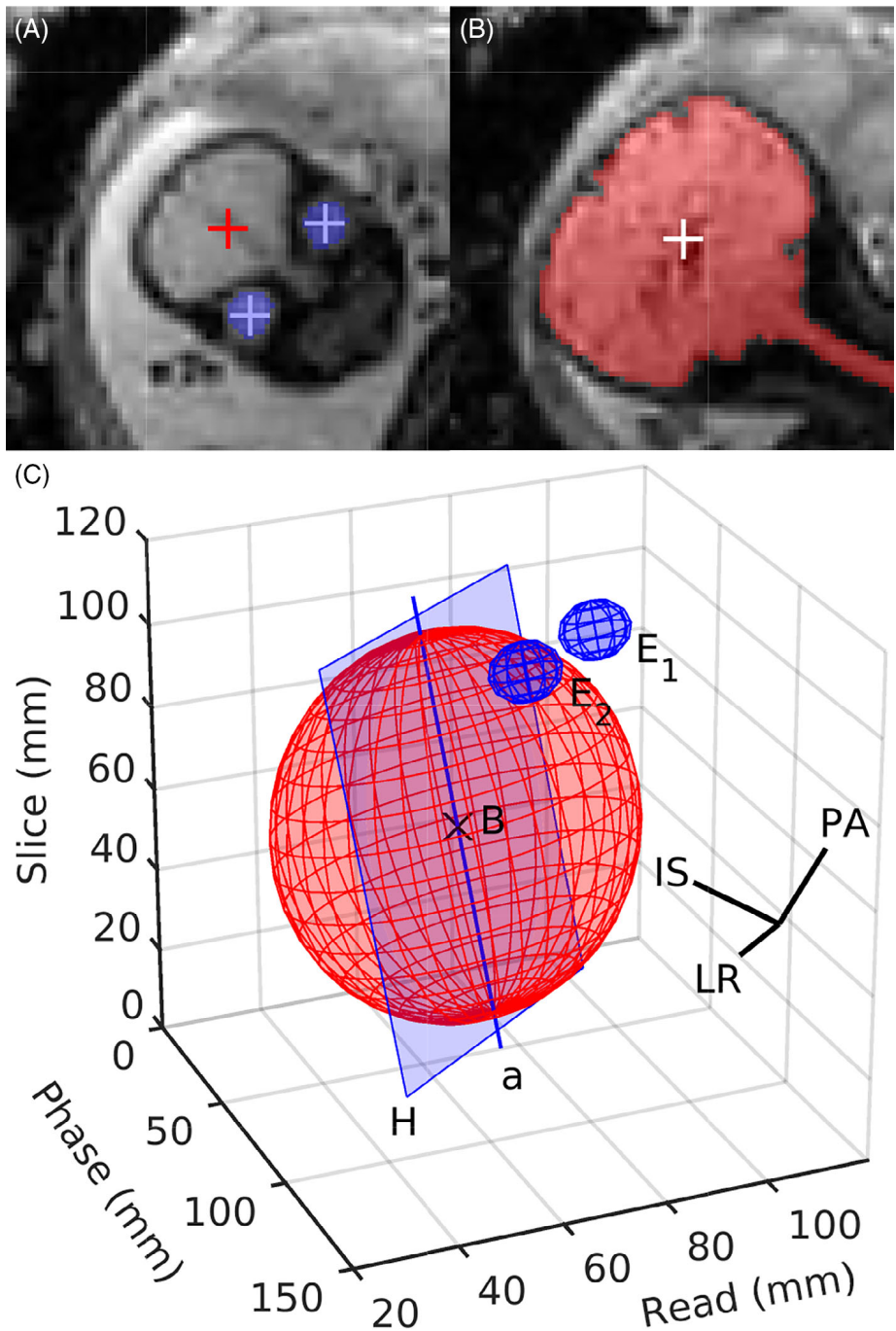

**FIGURE 6** Determination of the anatomical axes. (A) The location of the brain $B$ and eyes $E_{1,2}$, as indicated by the red and blue crosses, respectively, is combined with (B) geometric properties of the brain mask to derive the orientation of the brain in space. (C) The correct orientation of the inferior–superior axis IS and left–right axis LR is inferred by fitting an ellipsoid to the brain mask. This allows the construction of the plane $H$ that contains the major ellipsoid axis $a$ and is parallel to LR. The ellipsoid section lying within the half space which includes the eyes represents the inferior portion of the brain

informed consent is obtained from each participant before the experiment in accordance with the institution's Human Research Committee. A typical image slice is shown in Figure 3A. A full stack of double-oblique slices through the brain can be seen in Video S1. All of the following validation experiments are performed after the fact, and the pipeline is tested on an offline computer using images extracted from our PACS.

To test the system performance in cases with abnormal brains and in younger fetuses, we also consider two sets of full-uterus EPI stacks routinely acquired during clinical fetal MRI sessions at our institution between April and August 2020. First, we consider 12 stacks from 6 fetuses in the third trimester (30–36 weeks GA, mean ± SD 31.8 ± 3.1) with mild to severe ventriculomegaly or increased prominence of brain cerebrospinal fluid spaces. These are the most common neurological abnormalities in the third trimester seen at our radiology department, amounting to 43% and 29% of all cases in the considered time frame, respectively (50% each in this subset). Second, we consider 15 stacks from an additional 6 fetuses with normal brains in the second trimester (14–21 weeks GA, mean ± SD 18.7 ± 2.7). For both datasets, the acquisition planes and sequence parameters are similar to the above except that the in-plane resolution is $(2\ \text{mm})^2$.

## 2.7.2 | Ground-truth and success rates

Ground-truth brain orientations are obtained based on the location of the left/right eye and brain as determined by a trained labeler using FreeView.[47]

Our criterion for deeming automated brain-pose detection "successful" is correct localization of the left and right eye (accurate brain localization being a necessary condition for detecting the eyes). Quantitatively, the barycenter of the left eye has to be located within a radius $\varrho = 1.2 \times$ OD from the corresponding ground-truth landmark, and similarly for the right eye.

The threshold 1.2 is determined by manual inspection of the images after resampling in the derived anatomical frame using MATLAB (MathWorks, Natick, MA), being sufficiently strict to prevent false positives while allowing some variability in localization, for example, due to inter-slice motion. Specifically, resampling based on landmarks in proximity to the underlying anatomy produces views of standard anatomical planes of the brain. In contrast, if at least one of the landmarks does not correspond to the anatomy, resampling results in oblique slices relative to the brain or the image does not include the brain.

## 2.7.3 | Assessment of accuracy

Brain-localization accuracy is assessed in terms of the Euclidean distance $\Delta T$ from the "ground-truth" brain center. We assess the accuracy of brain orientations as the magnitude of rotation $\Delta R$ from the ground-truth head pose, calculated by regarding the sets of points $\{B, E_1, E_2\}$ as triangles and registering these together.[48] Acquisitions for which the pose detection is deemed a "failure" are excluded (see above).

## 2.7.4 | Intra-scout motion

We estimate the level of fetal-head motion during EPI scout acquisitions to analyze its effect on the detection errors $\Delta T$ and $\Delta R$. This is done offline by rigidly registering the sub-volumes consisting of odd- and even-numbered EPI slices, respectively, using FLIRT/FSL.[14] We use normalized correlation as similarity metric and confine the registration to a spherical mask of diameter $d = $ OFD with smoothed edges, positioned on the ground-truth brain center. Motion is summarized as a scalar motion score $S$ by summing magnitude translations and rotations, converting rotations to displacements on the surface of the brain,[49] which we model as a sphere of diameter $d$.

## 2.7.5 | Model parameters

Head-pose estimation depends on several parameters. We restrict the geometry of detected image features so that they are compatible with prior knowledge of the fetal anatomy. We explore the space of anatomical parameters to assess the extent to which they affect model performance. Parameters are varied one at a time, keeping all others fixed.

First, we assess model performance as the proportion of EPI stacks for which the left and right eye are correctly discriminated and located, using the criterion of Section 2.7.2 (accurate brain localization being a necessary condition for detecting the eyes). Second, we evaluate how robustly the model resolves the left–right ambiguity in labeling the eyes by computing a second success rate that is agnostic to left–right labels, and by comparing it to the first.

## 3 | RESULTS

## 3.1 | Problem quantification in the clinic

The typical success rate of an MRI technologist prescribing HASTE slices on the scanner is 74.5% of 106 clinical HASTE stacks. The remaining 25.5% of labeled stacks have slice orientations that clearly do not correspond to any of the sagittal, coronal or axial planes of the brain because of fetal-pose changes between consecutive HASTE stacks. However, due to motion-related artifacts or early termination of the acquisition, a label could not be unambiguously assigned to $n = 21$ additional scans, suggesting that at least ~38% of fetal-brain HASTE acquisitions may have limited to no diagnostic value. We do not identify any relationship between GA and the prevalence of incorrect orientation.

We stress that in contrast to motion during the scan, we only regard prescription errors and pose changes between the prescription and the acquisition as grounds for failure. A radiologist might consider an even larger proportion of stacks undiagnostic.

A second trained labeler assessed the same 127 HASTE stacks to provide an estimate of inter-rater reliability. No label was assigned to $n = 20$ stacks, and 24.3% of the remaining scans were found to have clearly incorrect anatomical orientations. A total of 89 stacks were rated by both labelers. Of these, 89.7% were assigned the same label, indicating low inter-rater variability. We determine Krippendorff's alpha coefficient to be $\alpha = 0.69$ (1 indicating perfect reliability, 0 no reliability).

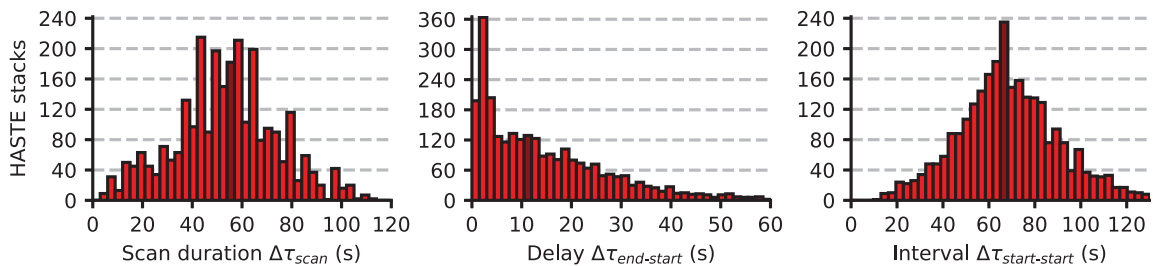

**FIGURE 7** Distribution of scan duration $\Delta\tau_{scan}$, delay between back-to-back scans $\Delta\tau_{end-start}$, and the resulting interval between contiguous scans $\Delta\tau_{start-start}$. We show data from 2722 fetal half-Fourier single-shot turbo spin-echo (HASTE) stacks acquired in 450 neurologically indicated sessions. The median interval is $\overline{\Delta\tau_{start-start}} = 67.0$ s, representing the average time between the last known head pose and the acquisition of slices based on that pose, during which head-pose changes may occur and cause incorrect anatomical orientation. We emphasize that the data shown include acquisitions that are interrupted early due to extreme subject motion. Dark red bars indicate the location of the median

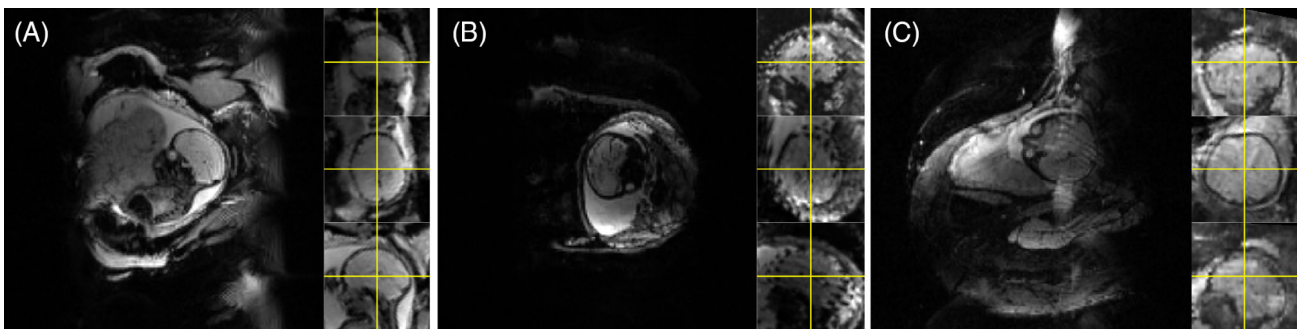

**FIGURE 8** Full-uterus echo planar imaging (EPI) scouts from three separate fetuses (29, 31 and 33 weeks GA) before and after resampling in the inferred anatomical frame. (A) Good-quality image acquired at low levels of motion. The brain is accurately located and aligned. (B) Successful orientation estimation despite motion artifacts. (C) Slice stack with wrapping artifacts (see "axial" view) resulting in a shifted brain-center estimate. This shift causes inaccurate orientation with the anatomical axes although the eyes are correctly identified. The views with the crosshairs correspond to coronal, axial and sagittal (140 mm)$^2$ portions extracted from the full (480 mm)$^2$ scout slices and illustrate how a subsequent half-Fourier single-shot turbo spin-echo (HASTE) scan could be automatically prescribed

Figure 7 shows the distribution of clinical scan durations and delays between back-to-back acquisitions for 2722 fetal HASTE stacks from 473 neurologically indicated sessions. The typical scan duration is $\overline{\Delta\tau_{scan}} = (54.4 \pm 20.9)$ s (median $\pm$ SD), and we emphasize that the data include scans interrupted due to subject motion. Most delays between contiguous acquisitions are short, with a median delay of $\overline{\Delta\tau_{end-start}} = (11.6 \pm 25.0)$ s. As a result, the typical interval between contiguous acquisitions is $\overline{\Delta\tau_{start-start}} = (67.0 \pm 30.1)$ s.

## 3.2 | Performance of the algorithm

The algorithm detects brain locations and orientations in agreement with the "ground truth" for 95.1% of EPI stacks from healthy fetuses in the third trimester, across a range of contrast, fetal age and motion levels. Video S2 shows a typical brain mask, obtained for the EPI stack of Video S1. The corresponding fetal-eye masks can be seen in Video S3. Figure 8 shows views extracted from typical full-uterus slice stacks of varying quality after resampling in the inferred anatomical frame.

The algorithm locates the brain in all except one EPI stack with artifacts, for which the subsequent local search of the eyes fails. In a second stack, the brain is found but its location is inaccurate due to motion, such that the algorithm misidentifies the eyes. In MATLAB, the detection including pre-processing takes (2.3 $\pm$ 0.3) seconds on average.

For fetuses in the third trimester with common neurological abnormalities, the performance remains robust. In 90.0% of EPI stacks, the system accurately locates the brain and eyes in the large FOV, and it correctly discriminates between the left and right eye in all but one stack. We leave out two of the stacks in this dataset because these are severely corrupted by maternal motion and half of the fetal brain is outside the FOV, respectively.

For younger fetuses, the performance is not optimal with only 18.2% of brain locations and orientations detected in agreement with the ground truth. In the remaining stacks, the detection fails as the system does not reliably localize the brain. We leave out two of the stacks in this dataset because most of the brain is outside the FOV. Based on this result, the following analyses focus on normal anatomy in the third trimester.

## 3.3 | Agreement with human labeler

Figure 9A shows the distribution of translational deviations $\Delta T_B$ ($\Delta T_{E1,2}$) from the "ground-truth" brain locations (eyes) in terms of Euclidean distance. These lie in the ranges $0.81$ mm $\leq \Delta T_B \leq 15.28$ mm ($0.22$ mm $\leq \Delta T_{E1,2} \leq 11.07$ mm). For the brain, the mean deviation is ~80% higher than for the eyes: $\overline{\Delta T_B} = (4.91 \pm 2.91)$ mm and $\overline{\Delta T_{E1,2}} = (2.73 \pm 2.02)$ mm (errors are SD).

The distribution of rotational deviations $\Delta R$ from the "ground-truth" head poses is shown in Figure 9B. Magnitudes of rotation are measured in the range $0.93° \leq \Delta R \leq 15.38°$, with mean $\overline{\Delta R} = (5.12 \pm 3.91)°$.

Table 2 summarizes the absolute translational mismatch $\Delta t_p$ and rotational mismatch $\Delta r_p$ for each axis of space ($p \in \{x, y, z\}$). The mean translational errors do not exceed $\overline{\Delta t_p} = 2.63$ mm along any axis, which is well below the voxel size. The mean rotational errors are below or equal to $\overline{\Delta r_p} = 2.70°$ for all axes. There is no identifiable relationship between fetal age GA and $\Delta T$ or $\Delta R$.

## 3.4 | Intra-scout motion

Fetal-head motion between odd- and even-numbered sub-volumes is estimated for 38 of 41 individual 2D-EPI stacks. Due to the short acquisition time, the amount of motion is generally low, with a few outliers: we measure summarized motion $S$ in the range $1.5 \leq S$ mm$^{-1} \leq 27.6$, with average motion of $\overline{S} = (8.5 \pm 7.9)$ mm (combined translations and rotations, error is SD). The maximum magnitude of translation (rotation) along (about) a single axis of space is $13.9$ mm ($13.4°$). There is no identifiable relationship between motion and the detection error $\Delta T$ or $\Delta R$.

For the remaining EPI stacks, the registration does not converge to an anatomically meaningful solution, as determined by visual inspection of the overlap of the brain and eyes between the sub-volumes. Two of these stacks exhibit large displacements between the sub-volumes (the algorithm successfully estimates the head pose from odd or even slices only). In the third EPI stack, the brain is partially out-of-view (causing the head-pose detection to fail).

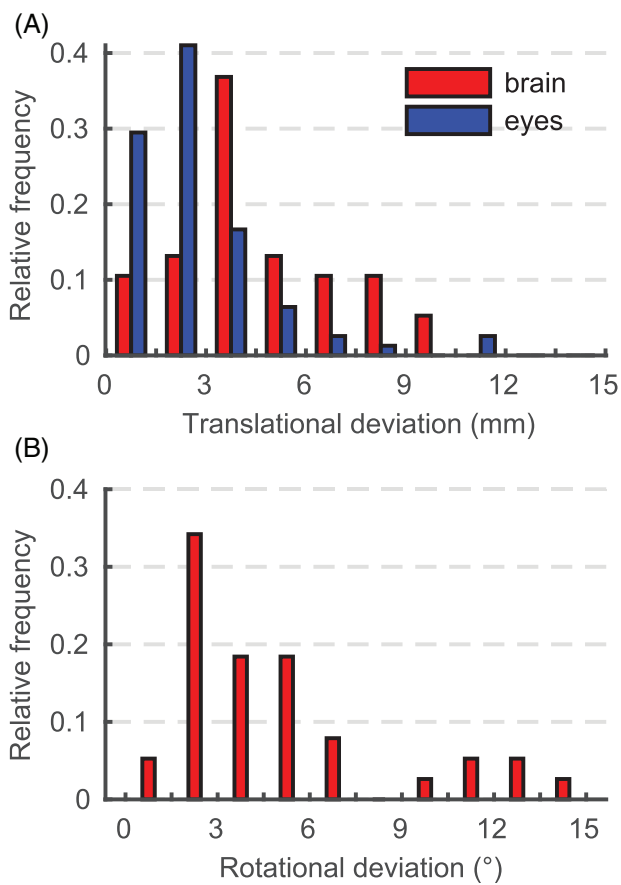

**FIGURE 9** (A) Distribution of deviations from the "ground-truth" landmarks in terms of Euclidean distance. (B) Distribution of the rotational deviation $\Delta R$ from the "ground-truth" head poses. Considered are only echo planar imaging (EPI) stacks for which landmarks correspond to the true anatomical locations

**TABLE 2** Deviation from the ground-truth brain location and orientation for each axis $p$ of space ($p \in \{x, y, z\}$)

| Statistic | Translation (mm) | | | Rotation (°) | | |
| --- | --- | --- | --- | --- | --- | --- |
| | $\Delta t_x$ | $\Delta t_y$ | $\Delta t_z$ | $\Delta r_x$ | $\Delta r_y$ | $\Delta r_z$ |
| Minimum | 0.08 | 0.15 | 0.00 | 0.03 | 0.14 | 0.01 |
| Maximum | 14.43 | 8.18 | 8.51 | 14.79 | 8.38 | 12.31 |
| Mean | 2.36 | 2.21 | 2.63 | 2.70 | 2.37 | 2.64 |
| SD | 2.56 | 1.99 | 2.20 | 3.12 | 2.27 | 2.70 |

*Note:* Statistics are shown for absolute translations $\Delta t_p$ and rotations $\Delta r_p$. Considered are only echo planar imaging stacks for which landmarks correspond to the true anatomical locations, in the third trimester and with normal anatomy.

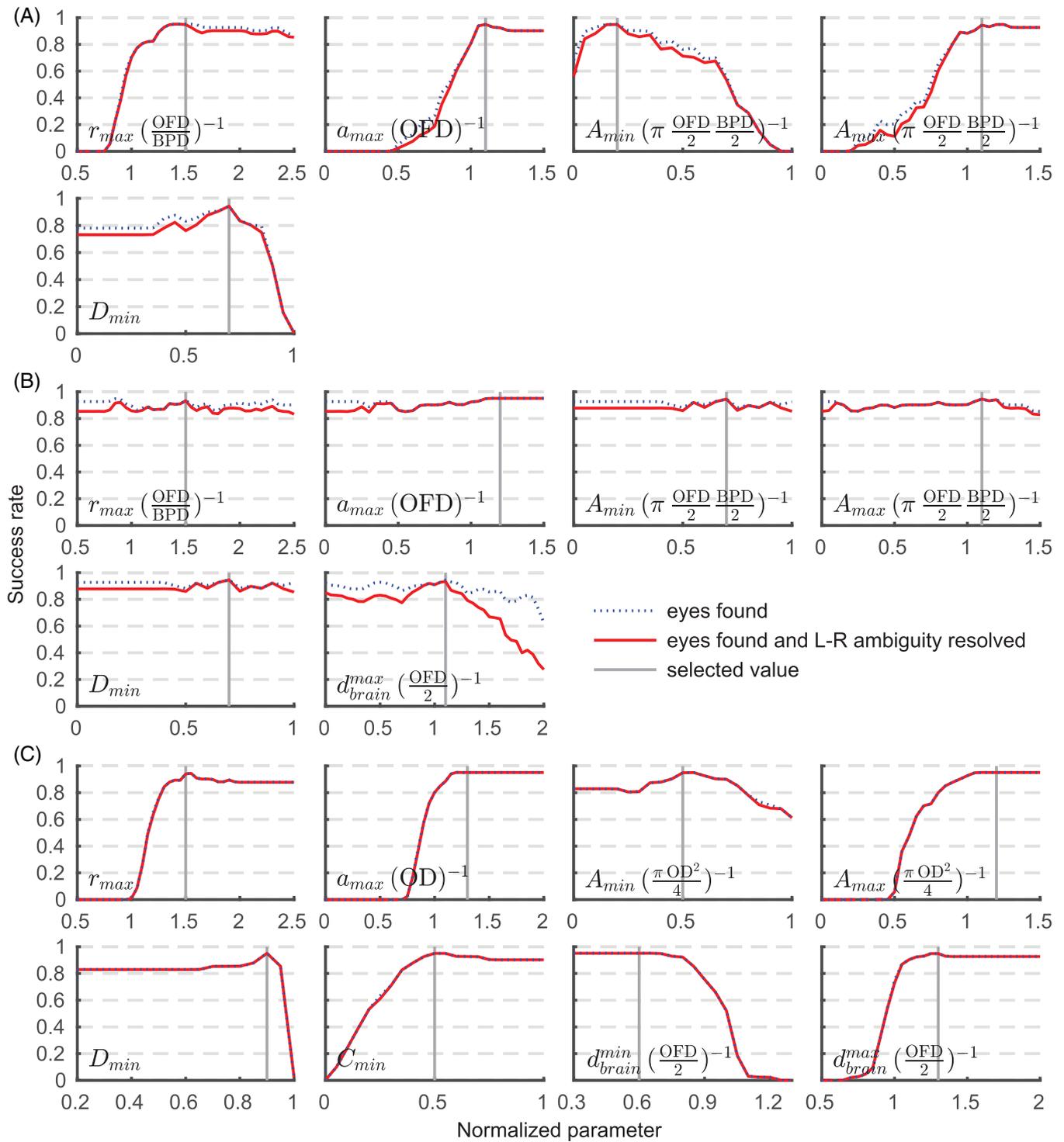

**FIGURE 10** Relationship of anatomical parameters from Table 1 on detection performance. Shown are parameters for (A) brain localization, (B) masking and (C) eye detection with units in parentheses. Some constraints of (B) have little impact on the success rate but boost efficiency by reducing the number of maximally stable extremal regions (MSERs) to process. Note that while the aspect ratio $r$ is defined as the proportion of major to minor MSER axes and therefore cannot be less than $r = 1$, it is given in units of occipitofrontal diameter (OFD)/biparietal diameter (BPD) in panels (A) and (B). See Section 2 for a full description of the parameters

## 3.5 | Model parameters

Figure 10 shows landmark identification and head-pose detection performance as a function of model parameters from Table 1. Some of the anatomical constraints used during local feature detection in Figure 10B, such as the maximum MSER area $A$, have little impact on the success rate. However, including these parameters boosts the efficiency of the algorithm by substantially reducing the number of MSERs to process, typically from several thousand to a few hundred at most.

## 4 | DISCUSSION

We propose a method for rapid fetal-head pose detection for low-resolution EPI stacks, which can be acquired in a fraction of the time needed for HASTE and thus be used as a scout scan for automatic slice prescription.

### 4.1 | 2D versus 3D

The use of MSERs for segmentation of brain MRI is proposed by Donoser and Bischof, who extend the approach to 3D.[50] We compared 2D and 3D MSER detection in 3D EPI but find the 3D version of the pulse sequence to have limitations over its 2D counterpart in terms of MSER detection, presumably because motion during the entire acquisition time (TA) is averaged into the volume, slightly blurring the anatomy at similar voxel size and TA. This is exacerbated by an increased TA, as the same FOV can be spanned with fewer excitations when a slice gap is used in 2D-EPI.

With 2D EPI, motion is effectively frozen during the ~30-ms acquisition of individual slices,[5] which leads to crisper images and improved MSER detection. However, motion between slices can cause intensity differences due to spin-history effects, impeding the detection of 3D MSERs across slice boundaries. Consequently, we find 2D acquisitions and 2D MSER detection more favorable to brain localization.

### 4.2 | Sensitivity to artifacts

While motion has no impact on the detection of individual MSERs, it can displace MSERs on adjacent slices relative to one another (interleaved slice ordering). This results in a shift between odd and even slices, leaving the center of the eyes and thus the head pose ill-defined. When the shift is smaller than the eyes, our algorithm tends to incorporate information from all slices into an average head-pose. This can be seen in Figure 8B, which shows an EPI stack with odd–even shift.

We observe that larger shifts typically cause the algorithm to consider odd or even slices only, selecting the pair of eye candidates with the lowest error $\epsilon$. Although only approximate estimates are needed to prescribe anatomical planes, this effect is undesirable and could be reduced by accelerating the scout acquisition (see below).

In contrast to motion, wrap-around artifacts caused by anatomy protruding from the FOV have a detrimental effect on the detection of MSERs (see Figure 8C). In 2D EPI, wrap is limited to the phase-encode direction and can be avoided by oversampling.

### 4.3 | Potential for online use

We showed that the median interval between contiguous fetal HASTE scans in the clinic is $\overline{\Delta\tau_{start-start}} = 67.0$ s. This represents the average time between the last known head pose and the acquisition of slices based on that pose. Fetal-pose changes can occur at any point during this time, causing incorrect anatomical orientation of the HASTE slices. The probability of pose changes occurring is substantially reduced if the brain orientation is inferred from a fast scout instead of the preceding HASTE stack. In fact, the average TA of our 2D-EPI stacks is only ~5 s, and our single-threaded algorithm takes ~2 s on an Intel Core 2.8-GHz i7-7700HQ CPU (Intel, Santa Clara, CA). This compares favorably to the interval $\overline{\Delta\tau_{start-start}}$ and has the potential to reduce vulnerability to head-pose changes when deployed for online slice prescription.

### 4.4 | Scout optimization

The EPI scout uses GRAPPA acceleration,[51] which requires an initial ~10-s calibration scan. This calibration needs to be acquired once at the beginning of the session and does not necessarily need to be repeated for subsequent scouts.[52,53] Recent methods acquire the autocalibration signal in under 2 s[54] using FLASH.[55]

The scout scan could be shortened by acquiring fewer slices with larger gap (at constant FOV). The large FOV is essential for localizing the fetal brain without prior knowledge of its location. At 31 weeks GA, the median OD is 14 mm.[46] Thus, increasing the gap from 50% to, for example, 75% would reduce TA by ~14% while leaving room for motion. Further speed-ups might include simultaneous-multi-slice imaging.[56,57]

We explored shortening TA by imaging at 4-mm resolution. However, the larger voxel size hampers the detection of MSERs, likely due to partial-volume effects.[58] Further work may realize speeds-ups using two specialized scouts: one low-resolution wide-FOV scan for brain localization, followed by a localized high-resolution scout for segmentation and eye detection. Using multiple MR contrasts might help with this hybrid approach.

## 4.5 | Speed versus accuracy

The pipeline is optimized to balance speed and accuracy since only coarse orientation with the anatomical axes is needed for the task of slice prescription. Any total execution time (scout acquisition plus orientation detection) below the current ~70-s delay introduced by visually inferring the brain orientation from the previous HASTE acquisition represents an improvement, reducing the time during which fetal-pose changes can occur. Therefore, the maximum acceptable runtime is 70 s, although the proposed pipeline is substantially faster (see Section 4.3).

In the third trimester, the algorithm robustly locates the fetal brain and eyes in over 94% of scouts, across a range of EPI contrasts, motion levels and abnormalities. Occasional detection failures caused by subject motion may not be problematic since the procedure is short, such that the scout and estimation could be repeated in practice. Further work may estimate the confidence in the head-pose detection, which would be informative if the pipeline is used on the scanner.

The mean deviation from ground-truth eye locations is just below the voxel size. The mean error on brain locations exceeds the value for the eyes by ~80%, resulting in an average head-pose discrepancy of 5.1° as compared to the ground truth. Larger errors for the brain are likely due to it being a larger structure and less spherical than the eyes. Manual localization from oblique 2D views is challenging, especially if the brain center is ill-defined due to motion-induced shifts between odd and even slices. Thus, the manually obtained landmarks cannot be considered absolute ground truth.

The precision of all measures considered is less than or equivalent to the voxel size. While there is good agreement overall, accuracies lower than what image registration achieves[59] were to be expected: instead of considering some $64 \times 64 \times 32$ voxels of typical EPI, the pose is derived from three points only.

While the system performs robustly in fetuses with common brain abnormalities, it is less robust in younger fetuses. The dark appearance of the skull is less prominent in the second compared to the third trimester, and the contrast between the brain and the surrounding fluids and tissues is reduced. These factors prove unfavorable to the MSER detection, causing the brain detection to fail in several images, such that the subsequent local search for the eyes is inaccurately initialized.

## 4.6 | Parameter space

We demonstrate accurate landmark detection in fetuses aged 26–37 weeks. Robustness to fetal development within the third trimester is achieved by filtering features commensurate with biometric data as a function of age. Although biometric information is available for 14–40 weeks GA,[44] further work is needed to extend the detection to the second trimester, for example, by acquiring a higher-resolution scout.

We show that the left–right ambiguity in identifying the eyes can be unequivocally resolved by carefully considering the shape of the brain mask. However, it may be unnecessary to do so for the purpose of slice prescription since the acquired images can be flipped along any axis at no cost and without introducing resampling artifacts. If this is the case, the masking procedure can be omitted to further reduce the runtime.

Some of the anatomical constraints in Table 1 may appear strict, for example, the surface area of the brain is not allowed to exceed the reference value by over 10%. We stress that these reference values are upper bounds since they are based on OFD, which is the largest diameter across the brain. In contrast, the algorithm operates in an arbitrary cross-section through the brain where principal MSER axes may be substantially lower than OFD.

Overall, we empirically choose conservative and slightly redundant restrictions to avoid mistakenly rejecting brain MSERs. Yet, if brain MSERs are rejected they are typically filled back in by the gap-closing procedure.

## 4.7 | Comparison with related work

Initially, we explored localization of the fetal brain based on template matching of the eyes.[18] However, we find template matching to take minutes and lack robustness in low-resolution EPI where the contrast between the eyes and surrounding tissues is not as high as in HASTE.

Keraudren et al[23] demonstrate brain localization in under a minute and brain extraction in another ~2 min for HASTE. These runtimes would be incompatible with on-scanner use, and the method does not localize the eyes, which we need for head-pose estimation. However, the idea of filtering MSER features by GA-dependent shape and size introduces robustness to age-related variability. We extend this approach to the fetal eyes and show that for our low-resolution EPI data, fast brain extraction is possible without resorting to classification.

Recent deep-learning methods perform head-pose estimation from high-resolution HASTE using convolutional encoder networks.[33,34] These models learn a function that can be evaluated in a fraction of a second, qualifying them for online use. However, they currently require preprocessing including brain extraction and N4 bias-field removal, the latter taking ~10 min per image.[35] In addition, these methods exploit custom hardware that is not necessarily available on the target scanner (see Section 1.3).

## 4.8 | Impact

We demonstrate that our pipeline can outperform a trained technologist in terms of speed when estimating the fetal-head pose from a fast EPI scout as opposed to the previous ~55-s HASTE acquisition. A trained technologist typically prescribes 75% of HASTE stacks with correct anatomical orientation, whereas the algorithm correctly estimates the head pose from over 94% of EPI scouts, for both normal and abnormal fetuses in the third trimester. However, the technologist has the disadvantage of possible changes in position during prescription, which is not a factor in evaluating the algorithm. Further work may determine if the success rate of the algorithm holds when assessing head poses in automatically prescribed HASTE.

There are several use cases for the technology. In addition to automated slice positioning, the MRI sequence may be automated such that the EPI scout and anatomical scan are alternated until views of the brain are obtained in all three planes. If automation helps reduce the time needed to acquire images with correct orientation, it will lower the cost of fetal MRI and increased subject comfort.

Our review indicates that at least 38% of acquisitions may be severely motion-corrupted. Thus, fetal MRI would benefit from prospective correction, that compensates for motion before data is acquired. Tisdall et al interleave volumetric navigators (vNavs)[60] with the anatomical sequence to measure head motion as it happens in the scanner. Fetal-brain tracking is demonstrated offline,[61] but further work is needed for vNav-based prospective correction. This contribution may be useful for automating the tracking on-scanner,[62] providing a brain mask that could guide the registration.

## 5 | CONCLUSION

In this paper we propose a method for fetal head-pose detection from fast, low-quality EPI scouts, with full control over all stages of the algorithm, enabling further optimization of individual components for real-time use on the scanner.

We present experiments showing that our algorithm robustly determines the fetal head-pose in over 94% of the slice stacks tested, across a range of EPI contrasts, motion levels and abnormalities in the third trimester. Robustness to fetal age is achieved using biometric data as a function of GA and demonstrated for fetuses aged 26–37 weeks. Our survey of clinical data suggests that at least 38% of fetal-brain scans are severely compromised by motion, indicating the need for automating the acquisition pipeline.

Operating on EPI instead of full ~55-s HASTE scans, our method has the potential to reduce the time during which fetal motion can corrupt the image geometry. When used for automatic slice prescription, the algorithm would remove the need for manually re-determining the orientation of the fetal brain from oblique views, which is challenging and time-consuming, and produce images available for direct radiological interpretation.

**ACKNOWLEDGMENTS**
This research was supported by NIH grants NICHD K99 HD101553, R01 HD085813, R00 HD074649, NINDS U01 AG052564 and NIBIB R01 EB008547. The authors thank Cindy Zhou and Elizabeth Holland for identifying relevant clinical data and coordinating scan sessions, and Hanna Loetz for support and encouragement.

**CONFLICT OF INTEREST**
Paul Wighton receives a salary from CorticoMetrics LLC for unrelated work.

**AUTHOR CONTRIBUTIONS**
**Malte Hoffmann:** Developed the method, worked out almost all the technical details, performed all numerical experiments, analyzed the data and designed the figures. **Malte Hoffmann:** Wrote and revised the article with input from all authors. **Esra Abaci Turk** and **Borjan Gagoski:** Carried out the in-vivo experiments and prepared data. **Leah Morgan:** Conducted quality ratings and discussed the rating procedure. **Paul Wighton:** Assisted in algorithm development and debugging. **M. Dylan Tisdall:** Devised the figure illustrating basis construction and encouraged Malte Hoffmann to investigate clinical scan timings. **Martin Reuter:** Assisted with image registration and discussed deep-learning methods. **Elfar Adalsteinsson** and **P. Ellen Grant:** Helped supervise the project and provided critical feedback that shaped the research. **P. Ellen Grant:** facilitated access to clinical data. **Lawrence L. Wald:** Proposed considering the fetal eyes for deriving the head pose. **Lawrence L. Wald** and **André J. W. van der Kouwe:** Conceived the original idea, supervised the project and guided Malte Hoffmann in key decisions. All authors discussed the results and commented on the manuscript.

**ORCID**

*Malte Hoffmann*  https://orcid.org/0000-0002-5511-0739
*Esra Abaci Turk*  https://orcid.org/0000-0003-0246-8793
*Borjan Gagoski*  https://orcid.org/0000-0003-3777-2621
*Leah Morgan*  https://orcid.org/0000-0001-6859-734X
*Paul Wighton*  https://orcid.org/0000-0002-6787-3856
*Matthew Dylan Tisdall*  https://orcid.org/0000-0002-0454-3112
*Martin Reuter*  https://orcid.org/0000-0002-2665-9693
*Patricia Ellen Grant*  https://orcid.org/0000-0003-1005-4013
*Lawrence L. Wald*  https://orcid.org/0000-0001-8278-6307
*André J. W. van der Kouwe*  https://orcid.org/0000-0002-2754-6594

## SUPPORTING INFORMATION

Additional supporting information may be found online in the Supporting Information section at the end of this article.